\documentclass[runningheads]{llncs}
\usepackage[T1]{fontenc}
\usepackage{cite}
\usepackage{hyperref}
\usepackage{subcaption}
\usepackage[algo2e,ruled,vlined,linesnumbered]{algorithm2e}
\usepackage{amsmath}
\usepackage{amssymb}
\usepackage{mathtools}
\usepackage{dsfont}
\usepackage{physics}
\usepackage[capitalize,noabbrev]{cleveref}
\usepackage{orcidlink}
\usepackage[misc]{ifsym}
\usepackage{booktabs} 
\usepackage{duckuments}
\usepackage{color}

\usepackage[font=small,skip=0pt]{caption}

\usepackage{tikz,pgf,pgfplots,pgfplotstable}
\pgfplotsset{compat=newest}
\usepgfplotslibrary{statistics}
\usepgfplotslibrary{groupplots}
\usepgfplotslibrary{fillbetween}
\input{tikz/colors_and_styles.sty}

\newcommand{\qt}{\textsc{QT}\xspace}
\newcommand{\kqt}{\textsc{KQT}\xspace}
\newcommand{\qtewma}{\textsc{QT-EWMA}\xspace}
\newcommand{\kqtewma}{\textsc{KQT-EWMA}\xspace}
\newcommand{\cpm}{\textsc{CPM}\xspace}
\newcommand{\spll}{\textsc{SPLL}\xspace}
\newcommand{\spllcpm}{\textsc{SPLL-CPM}\xspace}
\newcommand{\scanb}{\textsc{Scan-B}\xspace}
\DeclareRobustCommand{\vect}[1]{
  \ifcat#1\relax
    \boldsymbol{#1}
  \else
    \vb*{#1}
\fi}
\newcommand{\Break}{\textbf{break}}

\begin{document}
    \title{Change Detection in Multivariate data streams: Online Analysis with Kernel-QuantTree}
    %
    %
    \author{Olmo Nogara Notarianni\textsuperscript{(\Letter)} \orcidlink{0009-0001-1939-691X} \and
            Filippo Leveni \orcidlink{0009-0007-7745-5686} \and
            Diego Stucchi \orcidlink{0000-0002-8285-5285} \and
            Luca Frittoli \orcidlink{0000-0002-8205-4007} \and
            Giacomo Boracchi \orcidlink{0000-0002-1650-3054}}
    \author{Michelangelo Olmo Nogara Notarianni\inst{1}\textsuperscript{(\Letter)} \orcidlink{0009-0001-1939-691X} \and
           Filippo Leveni\inst{1} \orcidlink{0009-0007-7745-5686} \and
           Diego Stucchi\inst{2} \orcidlink{0000-0002-8285-5285} \and
           Luca Frittoli\inst{1} \orcidlink{0000-0002-8205-4007} \and
           Giacomo Boracchi\inst{1}\orcidlink{0000-0002-1650-3054}}
    \authorrunning{O. Nogara Notarianni et al.}
    \titlerunning{Online Analysis with Kernel-QuantTree}
    
    \institute{Politecnico di Milano (DEIB), Milan, Italy\\
               \email{\{michelangeloolmo.nogara, filippo.leveni, luca.frittoli, giacomo.boracchi\}@polimi.it}\\
               \and
               STMicroelectronics, Agrate Brianza, Italy\\
               \email{diego.stucchi@st.com}\\
    }
    \maketitle              
    %
    %
    \begin{abstract}
        We present Kernel-QuantTree Exponentially Weighted Moving Average (\kqtewma), a non-parametric change-detection algorithm that combines the Kernel-QuantTree (\kqt) histogram and the \textsc{EWMA} statistic to monitor multivariate data streams online. The resulting monitoring scheme is very \emph{flexible}, since histograms can be used to model any stationary distribution, and \emph{practical}, since the distribution of test statistics does not depend on the distribution of datastream in stationary conditions (non-parametric monitoring). \kqtewma enables controlling false alarms by operating at a pre-determined Average Run Length ($ARL_0$), which measures the expected number of stationary samples to be monitored before triggering a false alarm. The latter peculiarity is in contrast with most non-parametric change-detection tests, which rarely can control the $ARL_0$ a priori. Our experiments on synthetic and real-world datasets demonstrate that \kqtewma can control $ARL_0$ while achieving detection delays comparable to or lower than state-of-the-art methods designed to work in the same conditions.
    \end{abstract}

    \section{Introduction}
        \label{sec:introduction}

        Change detection is a frequently faced challenge in data stream analysis, where the properties of some monitored process, e.g. a measurement acquired by a sensor, may change over time. In machine learning, changes in data distribution are known as concept drifts and pose challenges for classifiers and learning systems in general, requiring continuous adaptation.
        
        In many application domains such as industrial monitoring, communication networks, and computer security, data come in virtually unlimited streams and need to be monitored \textit{online}. In particular, each new observation needs to be processed immediately after being acquired, and must be evaluated considering the whole data stream seen so far, while using a limited amount of memory and performing a fixed number of operations. In this study, we focus on online change-detection methods for multivariate data streams, which require algorithms capable of handling multidimensional vectors within these computational requirements and storage limitations.
        Another important challenge is posed by monitoring in a non-parametric manner, i.e., without any assumption on the initial data distribution. Non-parametric methods are particularly useful in real-world scenarios where the distribution of data is typically unknown. Unfortunately, most non-parametric change-detection algorithms are designed to monitor univariate data streams~\cite{CPM_statisticalProcessControl}.
        On top of that, controlling false alarms is a significant concern when each detected change can trigger costly interventions. Unfortunately, most online change-detection algorithms for multivariate data streams, particularly the non-parametric ones, struggle to effectively control false alarms. This paper addresses these challenges by proposing a method that is online, non-parametric, and capable of maintaining a target false alarm rate.

        Online change detection monitoring techniques can be grouped into two categories: \textit{one-shot} methods, which evaluate fixed-size batches of data points, and \textit{sequential} methods, which do not require a fixed sample size and take into account the whole data stream.
        QuantTree (\qt) is a \textit{one-shot} non-parametric solution which, supported by theoretical results, guarantees a pre-set constant false positive rate (FPR). First presented in~\cite{QuantTree}, \qt algorithm defines a histogram, partitioning the $d$-dimensional input space. Non-parametric statistics can be computed over it, enabling change detection in multivariate data streams batch-wise.
        A fundamental limitation of \qt is discussed in~\cite{KQT}: its splits are defined along the space axis, resulting in a hyper-rectangular partitioning that does not always adhere to the input distribution. To address this problem, a preprocessing stage is typically introduced to align the split directions to the principal components of the training set. However, it was observed~\cite{KQT} that this preprocessing can also worsen the control over false alarms. Hence, Kernel-QuantTree (\kqt) was introduced, a generalized version of \qt which partitions the space using kernel functions. The increased flexibility of the histogram in modeling the data distribution results in a \textit{one-shot} monitoring of multivariate data streams with increased detection power. However, \kqt is a batch-wise monitoring scheme using fixed-size windows and it fails to leverage the knowledge of the entire data stream distribution, thereby hindering fast detection of changes in an online scenario.
        A \textit{sequential} version of \qt algorithm, \qtewma, was presented in~\cite{QTEWMA}. \qtewma computes the Exponentially Weighted Moving Average (EWMA) statistic on a \qt histogram, thus considering the entire data stream acquired up to the current time instant $t$ to monitor the data distribution. Moreover, \qtewma can control the average time elapsed before a false alarm is triggered ($ARL_0$). Although being a truly \textit{sequential} extension of \qt, \qtewma inherits the same weaknesses of its \textit{one-shot} counterpart, i.e the axis-parallel splits.
        
        We propose Kernel-QuantTree Exponentially Weighted Moving Average (\kqtewma), a novel \textit{sequential} non-parametric change-detection algorithm for multivariate data streams that extends \kqt to the online scenario, following the approach of \qtewma. The theoretical properties of \kqt guarantee that \kqtewma is completely non-parametric since the distribution of our statistic does not depend on the data distribution, hence the thresholds controlling the $ARL_0$ can be set a priori, as in \qtewma. These thresholds guarantee by design a constant false alarm probability over time, thus a fixed false alarm rate at any time instant during monitoring.
        
        Our extensive experimental analysis on both synthetic and real datasets shows that \kqtewma outperforms state-of-the-art existing methods, successfully extending \kqt properties to the online scenario. Specifically, \kqtewma achieves control over the $ARL_0$ at a lower detection delay compared to competitors. We will show that, by relying on a precise partition of the space, \kqtewma outperforms \qtewma in complex scenarios, e.g., when analyzing data from multimodal distributions.

    
    \section{Problem formulation}
        \label{sec:problem_formulation}
        
        We consider a virtually unlimited multivariate data stream $x_1, x_2, \ldots$ in $\mathbb{R}^d$ where data samples $x_t$ are i.i.d. realizations of a random variable with unknown distribution $\phi_0$. We define the \textit{change-point} $t = \tau$ as the unknown time instant when the distribution $\phi_0$ experiences a change to $\phi_1$, i.e.:
        \begin{equation}
            x_t \sim
            \begin{cases}
                \phi_0 \quad \textnormal{if} \; t < \tau\\
                \phi_1 \quad \textnormal{if} \; t \geq \tau.
            \end{cases}
        \end{equation}
        We further assume we are provided with a training set $TR$ of $N$ stationary realizations from $\phi_0$, which is used to fit a model $\hat{\phi}_0$.
        
        After estimating $\hat{\phi}_0$, online change-detection algorithms typically compute a statistic $T_t$ at each observation $x_t$, to assess whether the new sequence $\{x_1, \ldots, x_t\}$ contains a change point or not. The decision rule usually involves checking whether $T_t > h_t$, where $h_t$ is a given threshold. The detection time $t^*$ is identified as the earliest time instant when sufficient statistical evidence indicates a change in the distribution, i.e.:

        \begin{equation}
            t^* = \min\{t : T_t > h_t\}.
        \end{equation}
        
        A desirable property of change-detection algorithms is that the sequence of thresholds $\{h_t\}_t$ can be set \textit{a priori} to guarantee a predefined $ARL_0 = \mathbb{E}_{\phi_0}[t^*]$, where the expectation is taken assuming that the whole data stream is drawn from $\phi_0$. In practice, $ARL_0$, represents the expected time before a false alarm occurs, and plays a role similar to Type I error probability control in hypothesis testing.
        The change-detection goal is to detect a distribution change as soon as possible, thereby minimizing the detection delay $t^* - \tau$, while aiming for an empirical $ARL_0$ to approximate the predefined target value set beforehand.

    \section{Related work}
        \label{sec:related_work}

        Change-detection algorithms are often \textit{parametric} since they underpin hypotheses about the data distribution $\phi_0$. As an example, the Change Point Model (\cpm)~\cite{CPM_statisticalProcessControl} based on the Hotelling test relies on the assumption that $\phi_0$ conforms to Gaussian distribution. \cpm performs online monitoring of data streams with theoretical guarantees regarding $ARL_0$ control. The \cpm formulation can be extended to detect changes given an unknown non-Gaussian distribution when implemented with the Lepage statistic~\cite{ross2011nonparametric}, always controlling false alarms. Unfortunately, the test statistics based on ranks are not suitable for multivariate scenarios. 
        A \textit{semi-parametric} change-detection strategy, Semi-Parametric Log-Likelihood (\spll), was presented in~\cite{SPLL}. \spll models the initial data distribution $\phi_0$ by fitting a Gaussian Mixture Model (GMM) $\hat{\phi}_0$ on a training set and then compares new incoming batches with batches from the training set using a likelihood test. Since \spll does not provide a way to set the detection threshold a priori to control the $ARL_0$, it was combined with \cpm\cite{QTEWMA}. In \spllcpm\cite{QTEWMA}, \spll reduces the dimensionality of incoming samples by computing their log-likelihood with respect to a GMM $\hat{\phi}_0$ fit on TR, thus the resulting univariate sequence can be conveniently monitored by a non-parametric extension of \cpm leveraging the Lepage test statistic~\cite{ross2011nonparametric}. Again, the main limitation of both \spll and \spllcpm is the assumption that $\phi_0$ can be well approximated by a probability distribution of a known family (a GMM), which does not hold in general.
        There are only a few other multivariate methods that perform \textit{non-parametric} change-detection. \scanb~\cite{li2018scan} employs a Maximum Mean Discrepancy (MMD) statistic and can be configured to achieve a target $ARL_0$. However, the thresholds for this method are deﬁned by analyzing the asymptotic behavior of $ARL_0$ when the size $B$ of the sliding window is large~\cite{li2018scan}.
        Therefore, it fails at accurately controlling $ARL_0$.
        The NEWMA algorithm~\cite{newma}, also based on MMD, examines the relationship between two EWMA statistics with distinct forgetting factors. A limitation of this approach is that setting the $ARL_0$ thresholds requires the known analytical expression of $\phi_0$. 
        The Kernel-CUSUM~\cite{kernelCUSUM} algorithm avoids assumptions about data distribution $\phi_0$, but relies on a truncated approximation for $ARL_0$, which results in the underestimation of the thresholds~\cite{kernelCUSUM}.
        \qt~\cite{QuantTree} and \qtewma~\cite{QTEWMA, QTEWMA_update} are histogram-based change-detection methods featuring the desirable property that the distribution of test statistics, defined over bin probabilities, does not depend on the initial distribution $\phi_0$. This allows to set detection thresholds a priori via Monte Carlo procedures,  allowing for efficient false alarm control.  
        \kqt~\cite{KQT} defines histogram bins via nonlinear partition of the input space, resulting in a powerful change-detection algorithm in multivariate data streams. However, being a \textit{one-shot} method, it cannot be directly employed for online change detection. Our proposal, namely \kqtewma, aims to extend \kqt to the \textit{sequential} scenario while retaining the capability of controlling false alarms given a target $ARL_0$ defined beforehand.

    \section{Kernel-QuantTree EWMA}
        \label{sec:kqt_ewma}

        We present \kqtewma, a novel change-detection algorithm which combines a \kqt histogram~\cite{KQT}, used as a model $\hat{\phi}_0$ of the stationary distribution $\phi_0$, and the online statistic $T_t$ based on an Exponential Weighted Moving Average~\cite{QTEWMA}. In Section~\ref{subsec:kqt_ewma_algorithm}, we illustrate the \kqtewma algorithm, describing the histogram construction, the threshold computation to control $ARL_0$, and the online monitoring process. In Section~\ref{subsec:computational_complexity}, we compare the computation complexity of \kqtewma against the alternatives considered in the experimental section. Finally, in Section~\ref{subsec:limitations} we discuss the limitations of \kqtewma.


        \subsection{The \kqtewma algorithm}
            \label{subsec:kqt_ewma_algorithm}

            
            Algorithm~\ref{alg:kqtewma} illustrates the training and inference phases of the \kqtewma algorithm.
            First, the \kqt histogram $h=\{(S_k,\pi_k)\}_{k=1}^K$ is constructed over the training set $TR\subset\mathbb{R}^d$ to match a set of target probabilities $\{\pi_j\}_{j=1}^K$ (line~\ref{alg:line:kqt}), as described in~\cite{KQT}. The histogram construction process consists in iteratively splitting the input space into $K$ bins defined as sub-level sets of a measurable function $\{f_k:\mathbb{R}^d\to\mathbb{R}\}_{k=1}^K$, which measures distances between training points and selected centroids using a kernel function. We remark that $K-1$ bins defined by \kqt are compact subsets of $\mathbb{R}^d$. A sample that does not fall into any of these is assigned to the \textit{residual} bin, which covers the unbounded remaining part of the space. 
            We consider the Mahalanobis and the Weighted Mahalanobis (WM) distances as kernel functions, (as in~\cite{KQT}), but the properties of \kqt hold for \textit{any} measurable function projecting a multivariate vector in $\mathbb{R}^d$ on a single dimension.

            As in~\cite{QTEWMA}, \kqtewma computes the weighted averages $\{Z_{j,t}\}$ (line~\ref{alg:line:Zt}), which keep track of the percentage of data stream samples $\{x_1,\dots,x_t\}$ falling in each bin $S_j$; to this purpose, we define $K$ binary statistics $\{y_{j, t}\}_j$ as:
            \begin{equation}
                \label{eq:QTEWMAy}
                y_{j, t} = \mathds{1}(x_t \in S_j),
            \end{equation}
            for each $j\in\{1,\dots,K\}$ and $t\geq 1$ (line~\ref{alg:line:y}).
            As discussed in~\cite{QTEWMA}, under the assumption that the monitored samples $x_t\sim\phi_0$ are stationary, the expected values of the binary statistics in \eqref{eq:QTEWMAy} can be approximated (line~\ref{alg:line:expected}) as:
            \begin{equation}
                \label{eq:QTEWMA_yexpected}
                \mathbb{E}[y_{j, t}] \approx \hat{\pi}_j := \frac{N \: \pi_j}{N+1}, \quad j<K \quad\; \textnormal{and} \quad\; \mathbb{E}[y_{K, t}] \approx \hat{\pi}_K := \frac{N \: \pi_K+1}{N+1}.
            \end{equation}
            
            During monitoring, each incoming sample $x_t$ is used to update the weighted averages $\{Z_{j,t}\}$ and to compute the test statistic $T_t$. First, the sample is processed by the histogram $h$ to obtain the binary statistics $\{y_{j,t}\}$ (line~\ref{alg:line:y}), which in turn are used to update the weighted averages $\{Z_{j,t}\}$ (line~\ref{alg:line:Zt}) as
            \begin{equation}
                \label{eq:QTEWMA_Z}
                Z_{j, t} = (1-\lambda) \: Z_{j, t-1}+\lambda \: y_{j, t} \quad\; \textnormal{where} \quad\; Z_{j, 0} = \hat{\pi}_j.
            \end{equation}
            The past samples are weighted by an exponential curve which decreases with time constant $\lambda$. The expected value of the $Z_{j, t}$ statistic under $\phi_0$ approximates $\hat{\pi}_j$, i.e. $\mathbb{E}[Z_{j, t}] \approx \hat{\pi}_j$ for $j = 1,...,K$, thus the change-detection statistic is computed (line~\ref{alg:line:Tt}) as follows:
            \begin{equation}
                \label{eq:QTEWMA_Tstatistic}
                T_t = \sum_{j=1}^K \frac{(Z_{j, t}-\hat{\pi}_j)^2}{\hat{\pi}_j}.
            \end{equation}
            The test statistic $T_t$ measures the overall difference between the proportion of points in each bin $S_j$, represented by $Z_{j, t}$, and their approximated expected values $\hat{\pi}_j$ under $\phi_0$, thus corresponds to the Pearson statistic. The statistic naturally increases as a consequence of a change $\phi_0 \rightarrow \phi_1$ that modifies the probability of at least one bin.
            Finally, the statistic $T_t$ is compared against the corresponding threshold $h_t$ to detect a change (line~\ref{alg:line:result}).

            \begin{figure}[t]
                \begin{algorithm2e}[H]
                    \caption{\kqtewma}
                    \label{alg:kqtewma}
                    \DontPrintSemicolon
                    \SetNoFillComment
                    \KwIn{training set $TR\subset\mathbb{R}^d$, target probabilities $\{\pi_j\}_{j=1}^K$, thresholds $\{h_t\}_t$, data stream to be monitored $x_1,x_2,\dots,x_t,\dots\subset\mathbb{R}^d$}
                    \KwOut{detection flag $CD$, detection time $t^*$}
                    Construct the \kqt histogram $\{S_j, \pi_j\}_{j=1}^K$ over $TR$ as in \cite{KQT} \label{alg:line:kqt} \\
                    Calculate the expected probabilities $\{\hat{\pi}_j\}_{j=1}^K$ as in \eqref{eq:QTEWMA_yexpected} \label{alg:line:expected} \\
                    Initialize the weighted averages $Z_{j, 0} \leftarrow \hat{\pi}_j$ for each bin $j\in\{1,\dots,K\}$ \label{alg:line:Z0} \\
                    Initialize the detection flag $CD \leftarrow$ False and the detection time $t^* \leftarrow \infty$ \label{alg:line:init} \\
                    \For{$t = 1 \dots$}{
                        Compute the binary mask $y_{j, t} \leftarrow \mathds{1}(x_t \in S_j)$ \label{alg:line:y} \\
                        Update the random variables $Z_{j, t} \leftarrow(1-\lambda) \: Z_{j, t-1} + \lambda \: y_{j, t}, \; \forall j = 1 \ldots, K$ \label{alg:line:Zt} \\
                        Compute the test statistic $T_t \leftarrow \sum_{j=1}^K(Z_{j, t}-\hat{\pi}_j)^2 / \hat{\pi}_j$ \label{alg:line:Tt} \\
                        \If{$T_t > h_t$}{
                            $CD \leftarrow$ True, $\; t^* \leftarrow t$ \label{alg:line:result} \\
                            \Break
                        }
                    }
                    \Return $CD$, $t^*$
                \end{algorithm2e}
            \end{figure}


        \subsubsection{False Alarms and Threshold computation strategy.}
        
        \kqtewma algorithm inherits from Kernel-QuantTree the fundamental property that the distribution of the statistics in~\eqref{eq:QTEWMA_Tstatistic} does not depend on the data distribution $\phi_0$. The true bin probabilities $p_j = \mathbb{P}_{\phi_0}(S_j)$, i.e. the set of probabilities of a point sampled from $\phi_0$ to belong to the bin $S_j$, are drawn from the Dirichlet distribution $(p_1, \dots, p_K) \sim \mathcal{D}(\pi_1N, \pi_2N, \dots, \pi_K N+1)$ where $\{\pi_j\}_{j=1}^K$ is the set of target probabilities, as demonstrated in \cite{KQT}. It follows that the distribution of any statistic based on \kqt, including $T_t$, does not depend on $\phi_0$~\cite{QuantTree, QTEWMA}.
        Thus, the thresholds $\{h_t\}_t$ can be defined a priori to control $ARL_0$ on any data stream, which is defined as:
            \begin{equation}
                \label{eq:QTEWMA_ARL0}
                ARL_0 = \mathbb{E}_{\phi_0}[t^*] = \frac{1}{\alpha}.
            \end{equation}
            
        As explained in~\cite{QTEWMA_update}, detection thresholds $h_t$  guarantee the constant false alarm probability, i.e. thresholds are such that the following equation is satisfied:
            \begin{equation}
            \label{eq:QTEWMA_thresholdsEq}
                \mathbb{P}(T_t > h_t \;|\; T_k \leq h_k \; \forall k < t) = \alpha \quad \forall t \geq 1.
            \end{equation}

           Since detection time $t^*$ is a Geometric random variable with parameter $\alpha$, the probability of encountering a false alarm before time $t$ can be determined through the geometric sum:
           \begin{equation}
           \label{eq:FArate}
               \mathbb{P}\left(t^* \leq t\right)=\sum_{k=1}^t \alpha(1-\alpha)^{k-1}=\alpha \cdot \frac{1-(1-\alpha)^t}{\alpha}=1-(1-\alpha)^t.
           \end{equation}
        Thus, we can monitor the control over false alarms in data streams containing a change point $\tau$ by computing the proportion of streams in which $t^* \leq \tau$.
      
        We leverage the results in~\cite{QTEWMA_update} which proves that, to estimate the thresholds $h_t$, one can directly simulate the construction of \qt histograms on a training set $TR \sim \phi_0$ of size $N$ by drawing its bin probabilities from the Dirichlet distribution, $(p_1, \dots, p_K) \sim \mathcal{D}(\pi_1N, \pi_2N, \dots, \pi_K N+1)$. This approach holds with \kqt given any kernel function, i.e. we can use the same threshold sequences given any measurable function, including linear split functions along the axis, which would result in a \qt histogram. Therefore, the same thresholds, computed in a Monte Carlo scheme, can be used for \qtewma and \kqtewma to guarantee a constant false alarm probability over time. The thresholds do not depend on the data distribution $\phi_0$ nor the data dimension $d$. The entire simulation procedure must be repeated when changing the $\lambda$ parameter of the EWMA statistic, the target bin probabilities $\{\pi_j\}_{j=1}^K$, or the training set size $N$.

        \subsection{Computational complexity}
            \label{subsec:computational_complexity}

            Since efficiency is key in online monitoring, we analyze the computational complexity of \kqtewma in comparison with \qt, \qtewma, \kqt, and \spll, \spllcpm, and \scanb. The results are summarized in Table~\ref{tab:complexities}. Further explanations can be found in~\cite{QTEWMA_update, KQT} 
            
            The training of a \kqt given a training set $TR$ of $N$ points comprises $i)$ the projection of $TR$ by $f_k$, whose cost depends on the specific kernel function, $ii)$ the computation of the split value, which costs $\mathcal{O}(N)$, and $iii)$ the centroid selection. The cost of computing the Euclidean distance - or other distances based on $l_p$ norms - is $\mathcal{O}(d)$, while the Mahalanobis distance costs $\mathcal{O}(d^2)$ and the Weighted Mahalanobis (WM) distance costs $\mathcal{O}(M \: d^2)$, where $M$ is the number of Gaussian components fitted to TR and $d$ is the data dimension. The centroid selection criteria is based on the information gain, which estimate is dominated by the computation of the determinant of the sample covariance matrix, which costs $\mathcal{O}(d^3)$. Overall, the cost of the index computation is multiplied by the number of centroids $V$ tested during the selection procedure; therefore, an upper bound for the cost of \kqt construction is $\mathcal{O}(K \: V \: (N + M \: N \: d^2+d^3))$ when using the WM distance and the information gain criteria. During monitoring, the only operation performed is the projection by $f_k$ of the samples, resulting in a cost of $\mathcal{O}(K \: M \: d^2)$ in case of the WM distance.

            \begin{table}
            \caption{
            Training and inference costs of \kqtewma with Weighted Mahalanobis (WM) distance and distances derived from $l_p$ norms (e.g. Euclidean distance when $p=2$), compared against the other considered methods.
            $V$ is the number of centroids tested to build each bin, $M$ is the number of Gaussian components fit on the dataset, $K$ is the number of bins, and $N$ is the training set size. As for the other methods, $m$ is the number of Gaussian components and $w$ is the window length used by \spll; $n$ is the number of windows of $B$ samples employed by \scanb.
            }
                \setlength{\tabcolsep}{4pt}
                \centering
                \begin{tabular}{lll}
                    \toprule
                    Method           & Training Cost                        & Inference Cost (per sample) \\
                    \midrule
                    \kqtewma (WM)  & $O(K \: V(N + M \: N \: d^2 + d^3))$ & $O(K \: M \: d^2)$          \\
                    \kqtewma ($l_p$) & $O(K \: V(N + N \: d + d^3))$        & $O(K \: d)$                 \\
                    \qtewma          & $O(K \: N \: \log N)$                & $O(K)$                      \\
                    \spll (online)   & $O(m \: N \: d^2)$                   & $O(m \: d + w \: \log w)$   \\
                    \scanb           & N.A.                                 & $O(n \: B \: d)$            \\
                    \bottomrule
                \end{tabular}
                \label{tab:complexities}
            \end{table}

            \subsection{Discussion and limitations} 
            \label{subsec:limitations}

            The main limitation of \kqtewma is that it is based on measures requiring the computation of the sample covariance matrix, which can be challenging in high-dimensional data streams. In \kqt, given any kernel function, the centroid selection criteria is the maximum information gain: the best split lowers the data entropy~\cite{KQT}, which is computed as $H(B)=(1 / 2) \log \left((2 \pi e)^d \operatorname{det}(\operatorname{cov}[B])\right)$, where $\operatorname{cov}[B]$ is the sample covariance matrix computed over a set of points $B$. Moreover, the sample covariance matrix estimated from the training set TR is used to define the Mahalanobis and the WM distances. The problem of determining the minimal sample size $N$ that guarantees that the sample covariance matrix approximates the actual covariance matrix depends on the data distribution, as well explained in~\cite{covariancematrixestimate}. Our experiments shows that \kqtewma can lose control over $ARL_0$ when few training points are provided. 

            QT-EWMA-update Algorithm~\cite{QTEWMA_update} is an effective monitoring scheme when $N$ is relatively small, i.e. when there are a few training samples, as this estimates the bin probabilities incrementally as new observations are available, as long as no changes are detected. While an incremental variant of \kqtewma can be implemented, this would be impractical due to computational and memory requirements, as it would require re-computing covariance matrices and centroids (possibly in a high dimensional space) at each update.

    \section{Experiments}
        \label{sec:experiments}

        The goal of our experiments is to show that \kqtewma controls the false alarms while achieving state-of-the-art detection delays. 
        To do this, we will show empirical results obtained on both synthetic and real-world data streams. In \kqtewma, as in \qtewma~\cite{QTEWMA}, we set the number of bins to $K = 32$ and uniform target probabilities $\pi_j = 1/K$. The exponential decay of the EWMA statistic is given by a time constant $\lambda = 0.05$. To monitor with \qt and \spll we set the batch size $\nu = 32$ as in \cite{QTEWMA}, and we employ the original configuration of the \scanb algorithm~\cite{li2018scan} ($n=5$ windows of $B = 100$ samples), if not specified otherwise.
        We set window length $w = 1000$ for \spllcpm. The number of centroids tested to build each bin of \kqtewma is $V = 250$.

        \subsection{Datasets}
            \label{subsec:datasets}

        \textbf{Synthetic}: As in~\cite{QTEWMA_update}, we generate synthetic data streams in spaces of increasing dimension $d \in \{2, 4, 8, 16, 32, 64\}$. We use Gaussian distributions $\phi_0$ with a random covariance matrix, and then we define the post-change distribution $\phi_1 = \phi_0(Q+v)$ as a random roto-translation of $\phi_0$. The roto-translation parameters $Q$ and $v$ are generated using the CCM framework~\cite{CCM} to guarantee a fixed distance between the two distributions computed as the symmetric Kullback-Leibler divergence $sKL(\phi_0, \phi_1) \in \{0.5, 1, 1.5, 2, 2.5, 3\}$. We expand our analysis to datasets sampled from bi-modal and tri-modal Gaussians to show the benefits of the distribution estimation with a \kqt histogram for change detection. 

        \noindent\textbf{Real-world}: As in~\cite{QTEWMA}, we test seven multivariate classification datasets of varying dimensionality:  El Ni\~{n}o Southern Oscillation (“ni\~{n}o”, $d = 5$), Physicochemical Properties of Protein Ternary Structure (“protein”, $d = 9$), two of the Forest Covertype datasets (“spruce” and “lodgepole”, $d = 10$), Credit Card Fraud Detection (“credit”, $d = 28$), Sensorless Drive Diagnosis (“sensorless”, $d = 48$), and MiniBooNE particle identification (“particle”, $d = 50$). We preprocess these datasets from the UCI Machine Learning Repository~\cite{UCI} as in~\cite{QTEWMA}: "particle", "protein", "credit", and "sensorless" datasets are standardized with respect to the standard score, and we sum to each component of “sensorless”, “particle”, “spruce” and “lodgepole” imperceptible Gaussian noise to avoid repeated values, which harm the construction of \qt histograms. The distributions of these datasets are considered to be stationary~\cite{QTEWMA}. We randomly sample the data streams and introduce changes $\phi_0 \rightarrow \phi_1$ by shifting the each distribution by a random vector drawn from a $d$-dimensional Gaussian scaled by the total variance of the dataset. We show the analysis of UCI datasets as the average results obtained over all the datasets. 

        Our experiments also include the INSECTS dataset~\cite{insects}, which contains $d = 33$ attributes derived from the wing-beat frequency of various insects, captured via an optical sensor. This dataset, tailor-made for change-detection techniques, contains records under diverse environmental conditions impacting insect flight behaviors. We focus on the abrupt-change variant of this dataset, which includes five distinct distribution changes $\phi_0 \rightarrow \phi_1 \rightarrow \cdots \rightarrow \phi_5$. We sample data points from these distributions to build our training set TR and test data streams. 
        Results obtained over the five changes present in the INSECTS datasets are averaged and shown all together. 
 

        \subsection{Figures of merit}
            \label{subsec:figures_of_merit}

            \textbf{Empirical $\mathbf{ARL_0}$.} To assess whether \kqtewma and the other considered methods control the target $ARL_0$ (see~\eqref{eq:QTEWMA_ARL0}), we compute its empirical value as the average time before raising a false alarm on data streams we sample from $\phi_0$. Empirical $ARL_0$ values are measured on $4000$ data streams drawn from $\phi_0$, given a target $ARL_0$ taking values in \{500, 1000, 2000, 5000\}. We generate stationary data streams of length $L = 6 \cdot ARL_0$ - the corresponding probability to detect a false alarm in each sequence is thus $ \mathbb{P}(t^* \leq L) \approx 0.9975$, as in~\cite{QTEWMA}.

            \noindent\textbf{Detection Delay.} We evaluate the detection power of \kqtewma and the other considered methods by their detection delay, i.e., $ARL_1 = \mathbb{E}[t^* - \tau]$, where the expectation is taken assuming that a change point $\tau$ is present. Again, we set the target $ARL_0$ a priori, $ARL_0 \in \{500, 1000, 2000, 5000\}$. Results are averaged over $4000$ data streams of length $6 \cdot ARL_0$, each containing a change point at $\tau = 300$. This is a difference compared to the analysis in~\cite{QTEWMA, QTEWMA_update}, where the average detection delay is computed at any given target $ARL_0$ on sequences of fixed length. We use sequences of the same length to estimate detection delay and $ARL_0$ to achieve a fair comparison between these two quantities. 
            
            \noindent\textbf{False Alarm rate}.
            The False Alarms (FA) rate is computed as the number of alarms raised at some $t < \tau $, averaged over $4000$ experiments. 
            By setting the target $ARL_0$ to $\{500, 1000, 2000, 5000\}$, we expect the percentage of false alarms to be $\{45\%, 26\%, 14\%, 6\%\}$, respectively, as stated by~\eqref{eq:FArate}. The target FA rates are indicated in the plots by vertical dotted lines.

        \subsection{Results and Discussion}
            \label{subsec:results_and_discussion}

    \begin{figure}[t!]
        \centering
        \vspace{-0.2cm}
        \includegraphics[width=\textwidth]{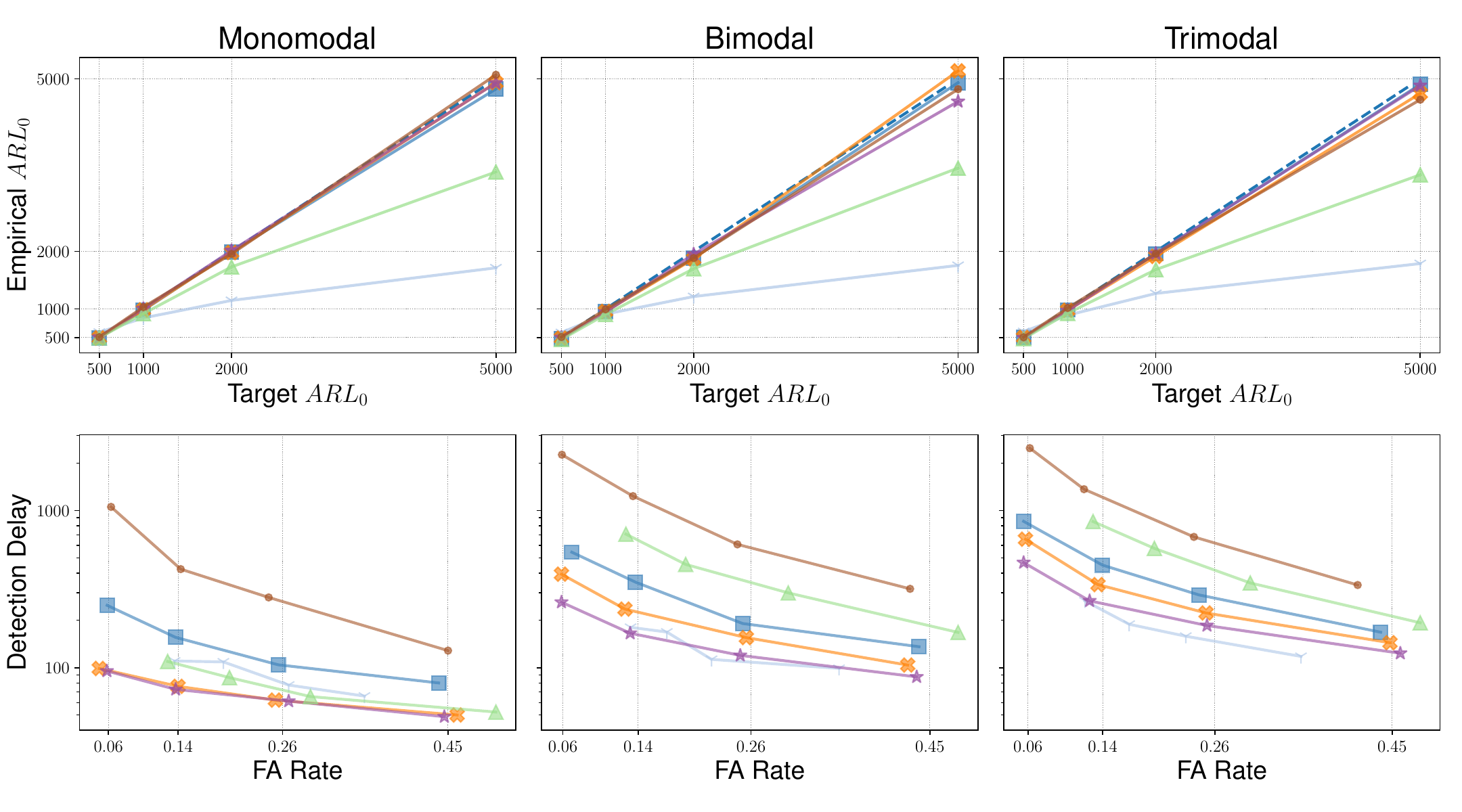}
        \vspace{0.1cm}        
        \includegraphics[width=0.7\textwidth]{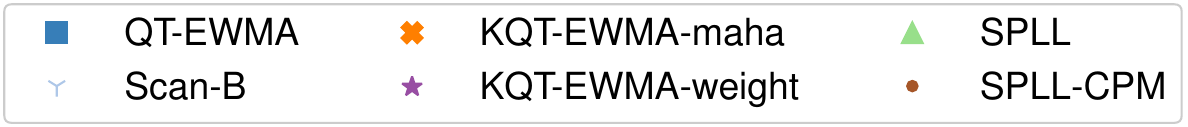}  
        \caption{
        Empirical $ARL_0$ and detection delay achieved by the considered methods monitoring data streams generated by Gaussian mixtures with increasing number of components ($1$, $2$, $3$). We show that as the number of components increases, \kqtewma with Weighted Mahalanobis (WM) distance advantage in terms of detection delay increases, achieving in general the lowest delays while controlling false alarms. In all the experiments, the GMM used to compute the WM distance fits $M=4$ components.
        }
        \label{fig:modalities}
    \end{figure}

    \begin{figure}[t!]
        \centering
        Results averaged over UCI+Credit datasets 
        \vspace{-0.2cm}
        \includegraphics[width=\textwidth]{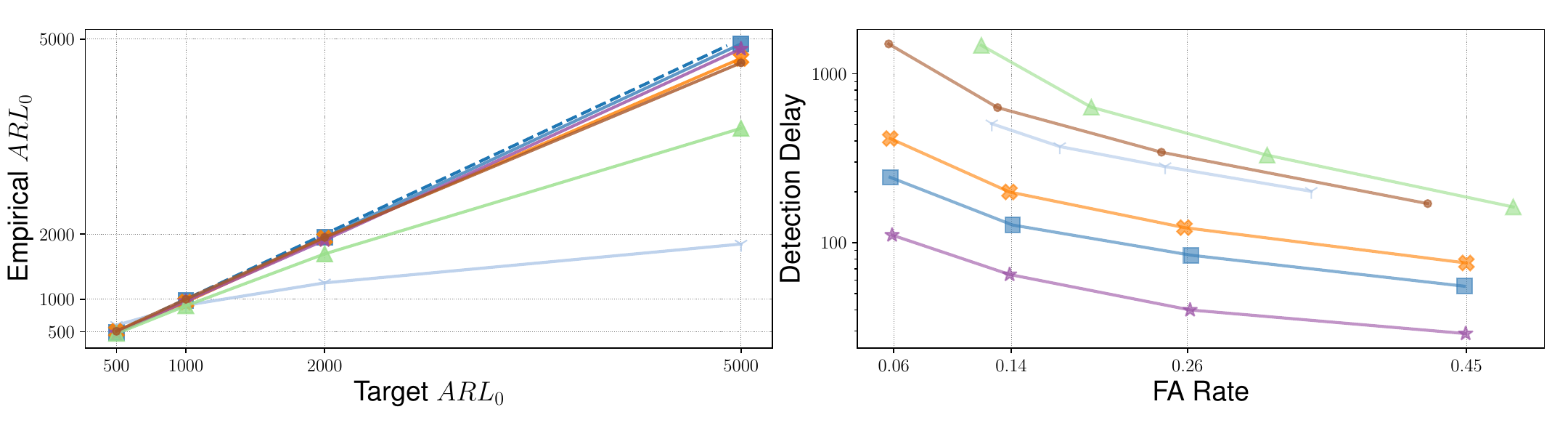}
        \includegraphics[width=0.7\textwidth]{pictures/legend2.pdf}  
        \vspace{0.1cm}
        \caption{
        Average empirical $ARL_0$ and detection delay on data streams sampled from the UCI datasets, excluding the highest-dimensional ones (i.e., "particle" and "sensorless"). In these two cases,  $N=4096$ training samples are not enough for \kqtewma based on Mahalanobis and WM distances to properly control $ARL_0$.
        In this setting, \kqtewma with WM distance achieves by far the best performance, halving the detection delay of \qtewma while controlling the target $ARL_0$.
        }
        \label{fig:UCIsmall}
    \end{figure}

    \begin{figure}[t!]
        \centering
        Results averaged over INSECTS dataset ($d=33$)
        \includegraphics[width=\textwidth]{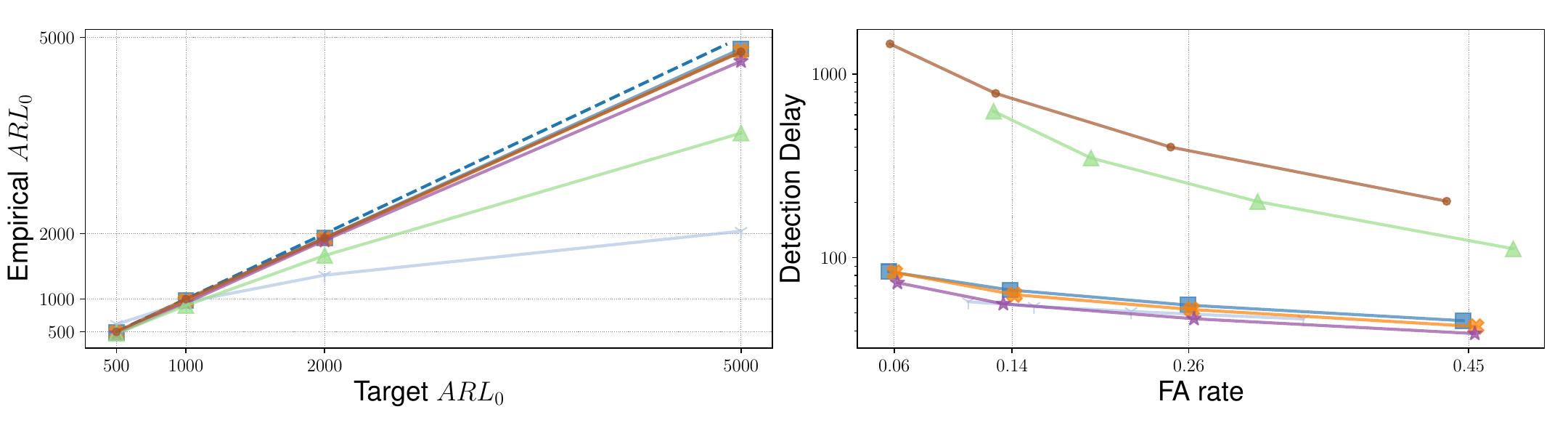}
        \caption{
        Average empirical $ARL_0$ and detection delay on data streams from the INSECTS dataset~\cite{insects}, with different combinations of $\phi_0$ and post-change distribution $\phi_1$. \qtewma and \kqtewma achieve similar detection delays, while \kqtewma with WM distance struggles in controlling higher values of $ARL_0$. 
        }
        \label{fig:insects_avg}
    \end{figure}

    \begin{figure}[t!]
        \centering
        Monomodal Gaussian ($d=4$)
        \includegraphics[width=\textwidth]{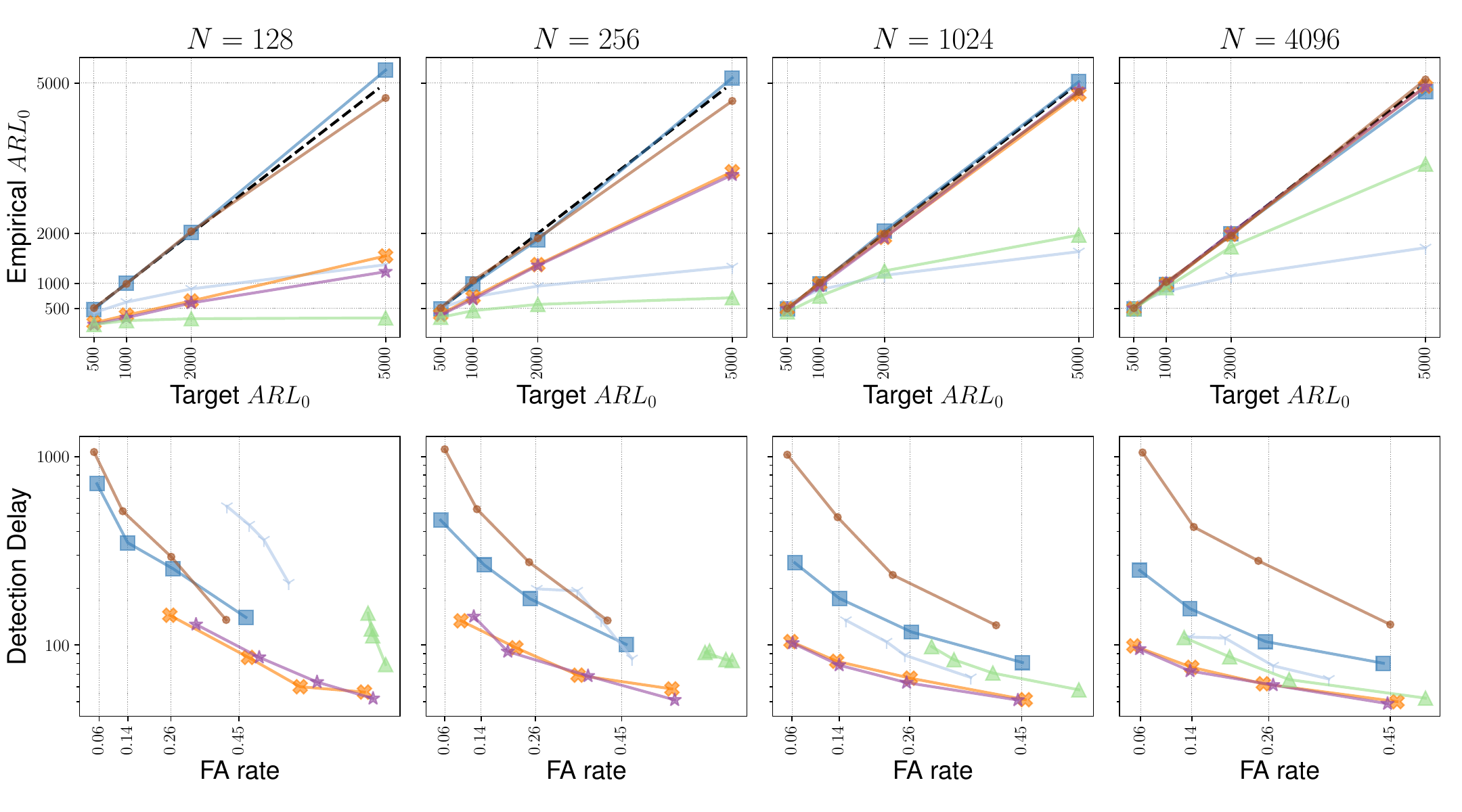}
        \includegraphics[width=0.7\textwidth]{pictures/legend2.pdf}  
        \vspace{0.1cm}
        \caption{Empirical $ARL_0$ and detection delay on Gaussian data streams in $d=4$ dimensions, for varying training set sizes $N\in\{128,256,1024,4096\}$. The empirical $ARL_0$ (first row) of \qtewma and \spllcpm always approaches the target values ($500, 1000, 2000, 5000$), while the other methods cannot control the $ARL_0$. When the training set size $N$ is sufficiently large ($N\in\{1024, 4096\}$), \kqtewma can control the FA rate, and achieves the lowest detection delay when using the Mahalanobis or the WM distance.}
        \label{fig:varyingN}
    \end{figure}


    \begin{figure}[t!]
        \centering
        Monomodal Gaussian ($N=4096$)
        \includegraphics[width=\textwidth]{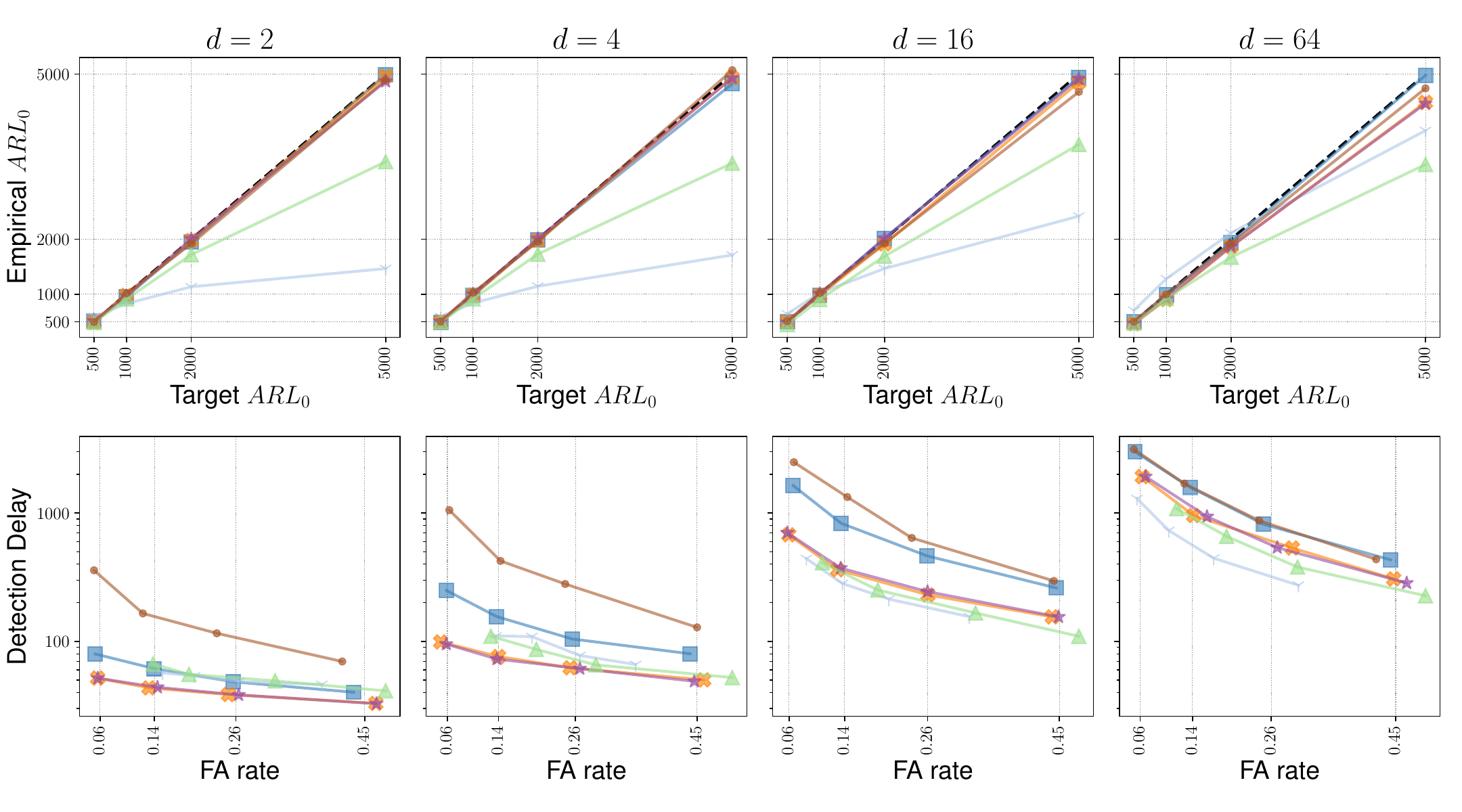}
        \includegraphics[width=0.7\textwidth]{pictures/legend2.pdf}  
        \vspace{0.1cm}
        \caption{
        Empirical $ARL_0$ and detection delay on data streams drawn from a Gaussian distribution with dimension $d\in\{2,4,16,64\}$, trained over $N=4096$ stationary samples.
        The empirical $ARL_0$ (first row) of \kqtewma, \qtewma, and \spllcpm always match the target, while \scanb and \spll fail.
        However, \kqtewma using the Mahalanobis distance cannot control the $ARL_0$ well when $d=64$, as $N=4096$ training points are not sufficient to estimate such a high-dimensional covariance matrix. When $d \leq 16$, the detection delay (second row) achieved with \kqtewma with Mahalanobis distance is the lowest achieved among the methods controlling the $ARL_0$. 
        }
        \label{fig:varyingDim}
    \end{figure}

    \begin{figure}[t!]
        \centering
        UCI "Sensorless" dataset ($d=48$)
        \includegraphics[width=\textwidth]{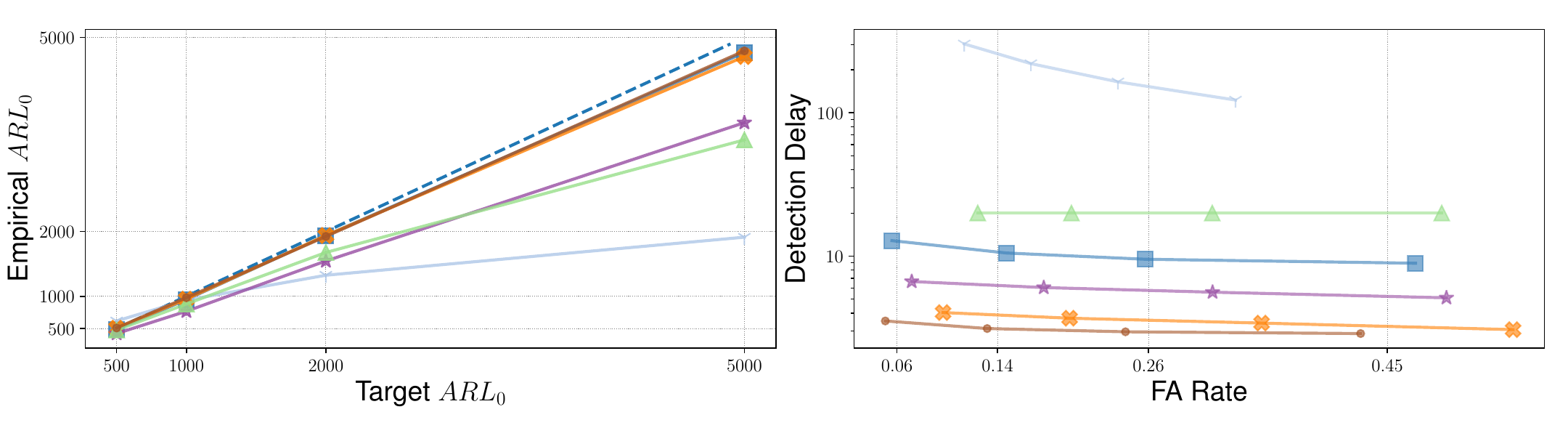}
        UCI "Particle" dataset ($d=50$)
        \includegraphics[width=\textwidth]{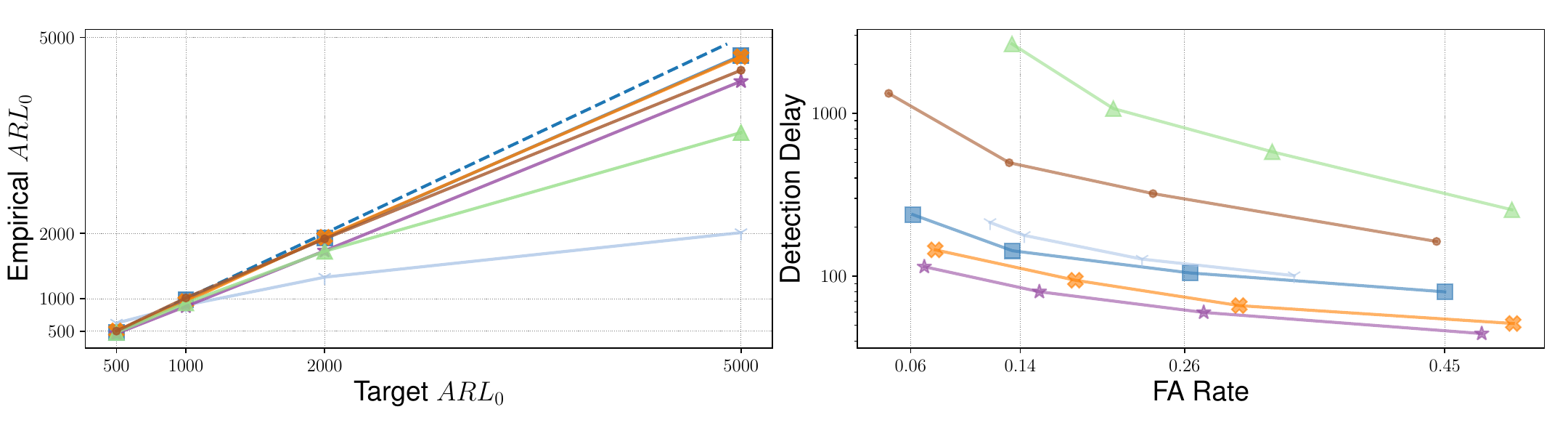}
        \includegraphics[width=0.7\textwidth]{pictures/legend2.pdf}  
        \caption{
        Empirical $ARL_0$ and detection delay achieved by the considered methods monitoring the two high-dimensional UCI data sets "particle" ($d=48$, above) and "sensorless" ($d=50$, below). The $N=4096$ training samples used in these experiments are not enough for \kqtewma based on Mahalanobis and WM distances to properly control $ARL_0$. Results are averaged over $4000$ experiments.
        }
        \label{fig:sensorlessParticle}
    \end{figure}

    \begin{figure}[t!]
        \centering
        Monomodal Gaussian ($d=4$, $N=4096$, Target $ARL_0=1000$)
        \includegraphics[width=0.8\textwidth]{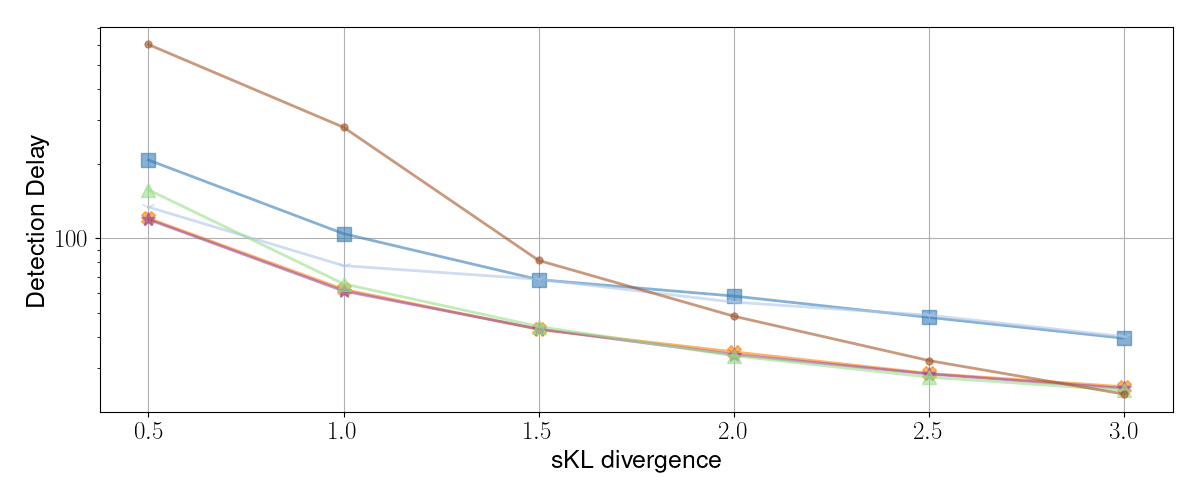}
        \includegraphics[width=0.7\textwidth]{pictures/legend2.pdf}  
        \caption{
        Detection delay as a function of the magnitude of the change $\phi_0\to\phi_1$ between pre- and post-change Gaussian sequences. We set the target $ARL_0=1000$.
        \kqtewma achieves the lowest detection delay, even in the challenging scenario when the change magnitude is low ($sKL=0.5$). As expected, all methods decrease their detection delays when the change magnitude increases. We remark that the empirical $ARL_0$ achieved by \spll and \scanb is lower than the target.
        }
        \label{fig:varyingsKL}
    \end{figure}

\noindent\textbf{False alarms control.}
To show the control over false alarms, we plot the empirical $ARL_0$ obtained in our experiments against the target $ARL_0$ set a priori. 
We compare results obtained over data sampled from monomodal Gaussians with the same performance measures computed over multimodal Gaussian datasets (bimodal and trimodal, in Figure~\ref{fig:modalities}, first row). In all the experiments, the number of components used to fit the GMM and compute the WM distance is $M=4$. \qtewma and \spllcpm can control all the target values chosen for $ARL_0$, while in general \spll and \scanb struggle in achieving high target $ARL_0 \in \{2000, 5000\}$. This is true also considering the results obtained with others real-world data sets (see Figures~\ref{fig:UCIsmall} and~\ref{fig:insects_avg}). 

\kqtewma effectively controls target values of $ARL_0$, while achieving the lowest detection delays in these scenarios(see  Figure~\ref{fig:modalities}, second row). 
Figure~\ref{fig:UCIsmall} (first row) shows experimental results averaged over data streams sampled from 5 UCI datasets having $d \leq 28$, for which we used $N=4096$ training points. We show two high-dimensional datasets (“particle” and “sensorless”, $d = 50$ and $d = 48$ respectively, see Figure~\ref{fig:sensorlessParticle}) separately, since $4096$ training points are not enough to estimate the sample covariance matrix in high dimensions, and \kqtewma based on Mahalanobis and WM distances do not properly control the $ARL_0$. In particular, \kqtewma with WM distance achieves low empirical $ARL_0$ values due to the difficulty in fitting a GMM here.

Figure~\ref{fig:varyingsKL} plots the change magnitude $sKL(\phi_0, \phi_1)$ against the detection delay achieved by the considered methods monitoring Gaussian data sequences with a target $ARL_0=1000$. While all methods successfully control the false alarms, \kqtewma achieves the lowest detection delay, even in this setting where the parametric assumptions of \spll are met (number of Gaussian components is set to $m=1$).
The advantage of \kqtewma over the alternatives is especially noticeable when the divergence between the pre- and post-change distributions is low ($sKL = 0.5$). As expected, the detection delay of all methods decreases when the change magnitude increases.

\noindent\textbf{Detection Delay vs False Alarms.}
To assess the detection power of these models, we plot the average detection delay against the percentage of false alarms.
Figure~\ref{fig:modalities} (second row) shows that \kqtewma with Mahalanobis and WM distances achieves the lowest detection delay regardless of the number of modalities, alongside \scanb. However, \scanb is unable to control higher values of $ARL_0$. Similarly, \spll cannot control $ARL_0$, and the distance from the target values is even more pronounced. This is also evident in the results on real data (Figures~\ref{fig:UCIsmall} and~\ref{fig:insects_avg}).
If we compare \kqtewma to \qtewma, which has the second-best detection delay values in the Gaussian scenario (Figure~\ref{fig:modalities}, second row), we can observe that \kqtewma more than halves the detection delay of \qtewma. Moreover, this difference increases as the complexity of the underlying distribution rises. This result is confirmed by all the experiments on both real (Figures~\ref{fig:UCIsmall} and~\ref{fig:insects_avg}) and synthetic (Figures~\ref{fig:varyingN} and~\ref{fig:varyingDim}) datasets, showing that the histogram construction strategy of \kqtewma, coupled with the Mahalanobis and WM distances, improves the detection performance, achieving lower detection delays. 
Figure~\ref{fig:varyingDim} shows the effects of \textit{detectability loss}~\cite{detectabilityLoss}: the ability to perceive a distribution change diminishes as the data dimensionality $d$ increases while the distance between pre- and post-change distribution is fixed ($sKL = 1)$, thus detection delays increase. 

Figure~\ref{fig:varyingsKL} illustrates the relation between the detection delay and the change magnitude between pre- and post-change Gaussian sequences. We set target $ARL_0=1000$ for all methods.
All methods successfully control the false alarms, and \kqtewma achieves the lowest detection delay, even when the parametric assumptions of \spll are met (number of Gaussian components is set to $m=1$).
The advantage of \kqtewma over the alternatives is especially noticeable when the divergence between the pre- and post-change distributions is low ($sKL = 0.5$). As expected, the detection delay of all methods decreases when the change magnitude increases.

Our extensive analysis shows that \kqtewma consistently achieves the lowest detection delays across different scenarios. Overall, \kqtewma based on Mahalanobis and WM distances can detect distribution changes more effectively, especially when the complexity of the underlying distribution rises.

    \section{Conclusion and Future Works}
        \label{sec:conclusion_and_future_works}
        We introduce \kqtewma, a non-parametric online change-detection algorithm for multivariate data streams based on Kernel-QuantTree~\cite{KQT}. The theoretical results underpinning \kqtewma~\cite{KQT, QTEWMA_update} guarantee the control of false alarms independently on the initial data distribution. Our experiments on synthetic and real-world data streams show that \kqtewma achieves state-of-the-art detection delay while effectively controlling false alarms. 
        
        In particular, the algorithm can leverage any measurable kernel function and it is able to fit complex distributions, resulting in high detection power. The sequences of thresholds can be computed independently on the data distribution $\phi_0$, the data dimension $d$, and the selected kernel function. Moreover, the monitoring scheme is invariant to roto-translation of the input data (when employing Mahalanobis and Weighted Mahalanobis distances, as shown in~\cite{KQT}), thus \kqtewma does not require any preprocessing step such as PCA.
        
        Our experimental evaluation also delineates some limitations: while the computational complexity of \qtewma scales well with the data dimension $d$ during both training and testing phases, \kqtewma's computational complexity does not, potentially impacting its practical utility in high-dimensional scenarios. Additionally, \kqtewma relies on the sample covariance matrix, whose estimation can be poor in high-dimensional scenarios where the training set TR is not sufficiently large. 
        Nevertheless, our experiments show that \kqtewma achieves excellent performance compared to the other methods designed for online monitoring, including \qtewma, while effectively controlling the false alarms, especially when considering complex data distributions such as multi-modal Gaussians or real-world datasets.

        Future work concerns addressing the limitations of \kqtewma with high-dimensional datasets. Specifically, we plan to design kernels that do not rely on covariance matrix computation and are specifically tailored for the \textit{sequential} high-throughput scenario.
        
        

        

    %
    %
    %
    \bibliographystyle{splncs04}
    \bibliography{bibliography}
\end{document}